\documentclass[final,12pt,authoryear]{elsarticle}
\usepackage{amssymb}
\usepackage{amsmath}
\usepackage{booktabs} % For formal tables
\usepackage{framed}
\usepackage{algorithm} 
\usepackage{algpseudocode} 
\usepackage{subcaption}

\journal{Arxiv}

\begin{document}

\begin{frontmatter}

% Edit the abstract: Mention AES is especially important for neuroevolution but is also general purpose and can be applied to other domains

% In related work: Add a section on synchronous parallel evolution and asynchronous evolution. Include a definition of asynchronous evolution. Mention the limitations of existing asynchronous algorithms and that they only work on a single population. 

% In related work: Add some new citations from 2021 to 2023 on asynchronous evolution.

% In the response to review #2, mention we are already comparing to state of the art (synchronous version of CoDeepNEAT)

% Update the pseudo code so that it uses Latex's pseudocode environment.

\title{Asynchronous Evolution of\\Deep Neural Network Architectures}

\author[a]{Jason Liang}
\author[a]{Hormoz Shahrzad}
\author[a, b]{Risto Miikkulainen}
\affiliation[a]{
    organization={Cognizant AI Labs},
    email={firstname.lastname@cognizant.com}
}
\affiliation[b]{
    organization={The University of Texas at Austin},
    email={risto@cs.utexas.edu}
}

\begin{abstract}

Many evolutionary algorithms (EAs) take advantage of parallel evaluation of candidates. However, if evaluation times vary significantly, many worker nodes (i.e.,\ compute clients) are idle much of the time, waiting for the next generation to be created. Evolutionary neural architecture search (ENAS), a class of EAs that optimizes the architecture and hyperparameters of deep neural networks, is particularly vulnerable to this issue. This paper proposes a generic asynchronous evaluation strategy (AES) that is then adapted to work with ENAS. AES increases throughput by maintaining a queue of up to $K$ individuals ready to be sent to the workers for evaluation and proceeding to the next generation as soon as $M<<K$ individuals have been evaluated. A suitable value for $M$ is determined experimentally, balancing diversity and efficiency. To showcase the generality and power of AES, it was first evaluated in eight-line sorting network design (a single-population optimization task with limited evaluation-time variability), achieving an over two-fold speedup. Next, it was evaluated in 11-bit multiplexer design (a single-population discovery task with extended variability), where a 14-fold speedup was observed. It was then scaled up to ENAS for image captioning (a multi-population open-ended-optimization task), resulting in an over two-fold speedup. In all problems, a multifold performance improvement was observed, suggesting that AES is a promising method for parallelizing the evolution of complex systems with long and variable evaluation times, such as those in ENAS.

\end{abstract}

\begin{keyword}
%% keywords here, in the form: keyword \sep keyword
Evolutionary Computation \sep Parallelization \sep Asynchronous Evolution \sep Sorting Networks \sep Multiplexer Design \sep Neural Architecture Search \sep Neuroevolution
%% PACS codes here, in the form: \PACS code \sep code

%% MSC codes here, in the form: \MSC code \sep code
%% or \MSC[2008] code \sep code (2000 is the default)

\end{keyword}

\end{frontmatter}

%% \linenumbers
%% main text

%% The Appendices part is started with the command \appendix;
%% appendix sections are then done as normal sections
%% \appendix

\section{Introduction}
\label{sec:introduction}

Evolutionary algorithms (EAs) have recently been extended to solving computationally expensive problems, such as evolutionary neural architecture search \citep[ENAS;][]{lu2018nsga,miikkulainen2019evolving,real2019regularized}. A main challenge in this domain is to reduce the amount of time spent in evaluating candidate solutions. For instance, when evolving architectures for deep neural networks (DNNs), a fitness evaluation includes training the network for multiple epochs, which is very time-consuming.

Fortunately, such evolutionary applications can take good advantage of parallel
supercomputing resources that have recently become available. Each evaluation can
be done on a separate machine (i.e.,\ a GPU resource), and thus the whole population can be evaluated at
the same time. However, individual evaluation times can vary significantly, making the process inefficient. A simple network may be trained in a few minutes, but larger ones may take several days on current GPUs \citep{miikkulainen2019evolving,liang:gecco19,liang:gecco21}. The EA has to wait for the longest evaluation to finish before the next generation can be created, during which time the other computational resources are idle \citep{scott:foga15}. As such, simple parallel EAs are not well suited for ENAS.

As a solution, this paper proposes an asynchronous evaluation strategy called AES that is designed to take full advantage of the available computational resources.
At each generation, a constant number of $K$ individuals are either being evaluated on the $R$ compute workers, have just finished their evaluation, or are waiting in a queue to be evaluated. As compute workers become available, they pull candidates from the queue for evaluation. As soon as a predetermined batch size of $M$ evaluations finish, $M$ new individuals are generated and placed in the queue. In this manner, all available computational resources are used at all times. This process can be seen as a mix of generational and steady-state GAs \citep{goswami2023variants}: Each batch of $M$ new individuals can be seen as a generation, but individuals from several generations may be evaluated in parallel.

AES was evaluated in a series of three experiments in this paper. The first experiment showed that in sorting network design (i.e.,\ a single-population optimization task with known optima), AES finds optimal solutions over twice as fast as synchronous evolution. However, evaluation times do not vary much in this domain; to demonstrate what is possible with higher variation, AES was applied to multiplexer design (i.e.,\ a single-population discovery task) with significantly more varied but still short enough evaluation times so that statistical significance could be measured. Further, the proper batch size $M$ was determined to be roughly 1/4 of the total population. The third experiment then scaled up the approach to ENAS for image captioning (i.e.,\ to multiple populations in an open-ended optimization task). In single production-size runs, AES was found to develop solutions of similar quality over twice as fast as synchronous evolution, and to find better solutions overall. AES is thus a promising tool for scaling up evolutionary algorithms to parallel computing resources. 

The main contributions of the paper are:
\begin{itemize}
\item A new algorithm, AES, for asynchronous evolution. 
\item A demonstration of the effectiveness of AES over synchronous EAs experimentally on a typical EA as well as ENAS.
\item A statistical analysis of how and why AES is able to gain this performance advantage.
\end{itemize}

The rest of the paper is
organized as follows: Section~\ref{sec:background} reviews related work on parallel EAs, asynchronous EAs, and neuroevolution of DNNs.
Section~\ref{sec:algorithm} introduces the generic version of AES and describes how it can be adapted to ENAS. Section~\ref{sec:experiment}
presents experimental results comparing the performance of AES with
synchronous evaluation in the sorting-network, multiplexer, and ENAS domains. Section~\ref{sec:discussion} analyzes the sources of the speedup
and proposes future work.

\section{Related Work}
\label{sec:background}

This section reviews prior work on parallel evaluation strategies for EAs, asynchronous evaluation strategies, and methods for evolutionary neural architecture search.

\subsection{Parallel Evaluation Strategies}

A common computation bottleneck in EAs is the evaluation step, where all individuals in the population must have their fitness determined. To overcome this bottleneck, the evaluation can be performed in parallel on multiple computing resources simultaneously. The simplest strategy is to run copies of a separate independent EA on each worker node and to return the best result of all the independent runs \citep{sudholt2015parallel}. While this strategy works for EAs where evaluation times are relatively short, it does not scale to problems where evaluation time are extremely long, such as neuroevolution.

A more synergetic approach is global parallelization \citep{adamidis1998parallel, schuman2016parallel}, where a single master node (server) assigns individuals to multiple worker nodes for evaluation. When all individuals have been evaluated, the master node proceeds onto the next generation. Further, multiple master nodes can be linked to a super-master node, which design may improve performance especially with a large number of workers \citep{schuman2016parallel}. Unfortunately, when evaluation times vary significantly, the master node will have to wait for the slowest worker node, during which the other workers are idle. Global parallelization is thus poorly suited for neuroevolution, where evaluations are not only long, but also vary widely in length.

Another challenge with parallelizing EAs is the communication bottleneck that occurs when individuals are sent to workers. To overcome this challenge, an island model \citep{adamidis1998parallel,sudholt2015parallel} can be used, where the population is divided into several subpopulations and each subpopulation is evolved and evaluated on a separate worker node. Periodically, the worker nodes exchange individuals in order to avoid converging to local optima. The topology in which the worker nodes are linked to each other is a major design concern for island models. An even more fine-grained parallelization approach is the cellular model \citep{sudholt2015parallel}, where each worker node is assigned to a single individual. Island and cellular models are typically synchronous and still suffer from the same issues as other synchronous EAs. Luckily for most problem domains, including neuroevolution, communication between workers is not the bottleneck, but rather the computation time spent by the worker nodes themselves.

\subsection{Asynchronous Evaluation Strategies}

Asynchronous evaluation strategies are a way to overcome the issues with synchronous EAs. The key difference is that while synchronous EAs are based on a generational model, asynchronous EAs are a parallel version of steady state EAs \citep{scott2015evaluation,scott:foga15}. In other words, they do not wait for the entire population to be evaluated but proceed with evolution when only a subset of the population has been evaluated. As a result, asynchronous EAs are well suited for problems with long, highly variable evaluation times. This class of EAs were proposed in the early 1990s, and have been occasionally used by practitioners \citep{rasheed:cec99,depolli:ec13,luke:ecj14,harada2020analysis}. However, little work is done analyzing the behavior and benefits of such algorithms \citep[see][for exceptions]{zeiglerICGA93,kim:phd94,scott:foga15,abdelhafez2019performance}. Such methods have recently become more relevant when parameter tuning for large simulations has become more common \citep{kiran2022hyperparameter}.

A current state-of-the-art approach to asynchronous evaluation methods is SWEET \citep{scott2023avoiding}, which allows individuals to be selected as parents for reproduction even if these individuals have not finished evaluation. Since fitness information is not available, the parents are randomly selected as input into a tournament-selection operator.  Other recent approaches include TLAPEA \citep{harada2020study}, which automatically determines the optimal time to wait after the evaluation of an individual before continuing evolution. These methods are similar to AES in that both rely on partial evaluation of the population before continuing with evolution. However, while they improve over a basic asynchronous algorithm, they still favor solutions that evaluate fast. This limitation makes SWEET and TALAPEA less well suited for neuroevolution, where the solutions (i.e.\ network architectures) that take the longest to evaluate may perform the best.

An asynchronous neuroevolution algorithm that is closely related to the AES is rtNEAT
\citep{stanley:ieeetec05,papavasileiou2021systematic}. In that approach, a population of neural networks is
evaluated asynchronously one at a time. Each neural network is tested in a video
game, and its fitness is measured over a set time period. At the end of the period,
it is taken out of the game; if it evaluated well, it is mutated and crossed over
with another candidate to create an offspring that is then tested in the game. In
this manner, evolution and evaluation are happening continually at the same time.
The goal of rtNEAT is to make replacement less disruptive for the player; it was not designed to parallelize evolution to speed it up in a distributed environment. Thus, unlike AES, it does not provide a performance advantage \citep{stanley:ieeetec05}. 

The rtNEAT approach to asynchronous neuroevolution assumes that the evaluation times are approximately the same. Therefore, it is not well suited for neuroevolution where evaluation times can vary significantly. Such variation is especially prominent in the training of deep neural network architectures, described in the next section. Another drawback of existing approaches is that they are only designed for a single population of individuals \citep{scott:foga15,chitty2021partially}. As such, existing asynchronous EAs cannot deal with the coevolution of multiple populations, such as in the neuroevolution domain described later. AES was designed to overcome these limitations, and thus presents an improvement over rtNEAT and asynchronous EAs in general.

\subsection{Evolutionary Neural Architecture Search (ENAS)}

\begin{figure}[ht]
  \begin{center}
    \includegraphics[width=\columnwidth]{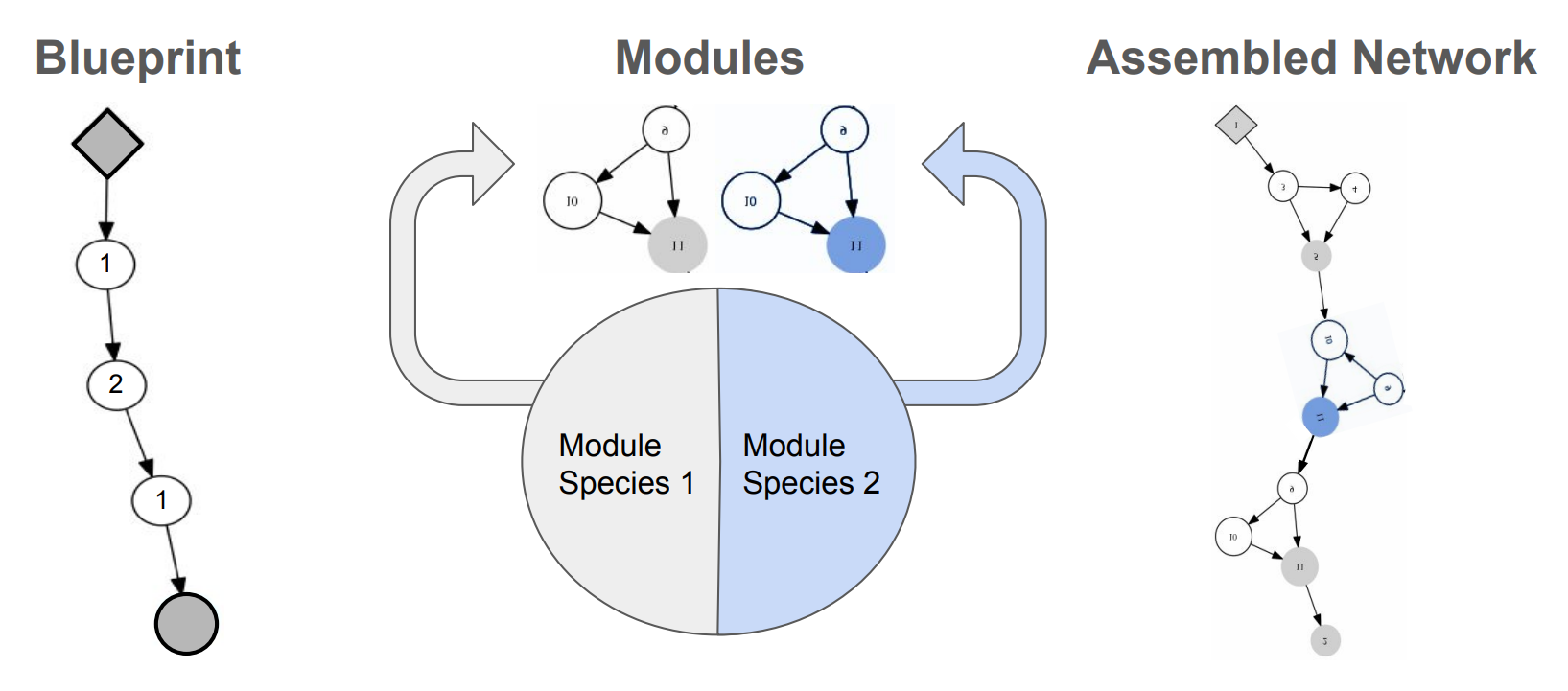}
    \caption{A visualization of how CoDeepNEAT assembles networks for
      fitness evaluation. The blueprints and modules are evolved in separate populations, divided into species (or subpopulations). For evaluation, they are assembled into a network by replacing the blueprint nodes
      with modules drawn from the corresponding module species. This approach makes it possible to evolve repetitive and deep structures seen in many recent successful DNNs.}
    \label{fg:assembling}
  \end{center}
\end{figure}

DNNs have achieved state-of-the-art performance on many machine learning
competitions and benchmarks in areas like computer vision, speech, and natural language processing \citep{collobert:icml08,graves:icassp13,szegedy:cvpr16,dosovitskiy2020image}. Often, hyperparameter choice and the structure of the network have a massive impact on its performance, and as a result, much research effort has been spent on discovering better
architectures \citep{he:arxiv16,szegedy:cvpr16,tan2019efficientnet,wu2021cvt}. 

Recently, EAs have been proposed as a viable way to optimize the architecture and hyperparameters of a DNN automatically \citep{miikkulainen2019evolving,liang:gecco18,liang:gecco19}. Evolution can generate DNNs with diverse topologies and achieve state-of-the-art performance on large-scale visual domains \citep{real2019regularized}. In addition, they can optimize multiple conflicting objectives such as performance and network complexity \citep{lu2018nsga}. Advanced EAs like CMA-ES \citep{loshchilov:arxiv16} can discover good hyperparameters in high-dimensional search spaces \citep{loshchilov2016cma}, performing comparably with statistical algorithms such as Bayesian optimization \citep{snoek2015scalable}. 

In this paper, a powerful EA called CoDeepNEAT \citep{miikkulainen2019evolving} 
is used to explore the search space for potential DNN topologies and hyperparameters.
CoDeepNEAT consists of a population of blueprints and a population of modules. Each
population is evolved separately with a modified version of NEAT
\citep{stanley:ec02}.  NEAT automatically divides each population into subpopulations, or species, of similar
individuals. An individual in the blueprint population is a graph
where each node contains a pointer to a particular module species in the module population.
An individual in a module population is a graph where each node represents a particular
DNN layer and its corresponding hyperparameters (number of neurons, activation
function, etc.). As shown in Figure~\ref{fg:assembling}, the modules and
blueprints are combined to create a temporary population of assembled networks. 

Each individual in this assembled population is then evaluated by training it
on some supervised task, determining their performance on a dataset, and
assigning that performance metric as fitness. The fitness of the individuals
(networks) is attributed back to blueprints and modules as the average fitness
of all the assembled networks containing that blueprint or module. One of the
advantages of CoDeepNEAT is that it can discover modular, repetitive
structures seen in state-of-the-art networks such as Googlenet and ResNet
\citep{szegedy:cvpr16,he:arxiv16}. CoDeepNEAT has achieved state-of-the-art performance in multiple problem domains, including image captioning, multitask learning, and automatic machine learning \citep{miikkulainen2019evolving,liang:gecco19,liang:gecco21}. However, CoDeepNEAT still suffers from the same issues as any synchronous EA.  Improving CoDeepNEAT through asynchronous evaluation of the population is the main technical challenge solved in this paper.

\section{The Asynchronous Evaluation Strategy Method (AES)}
\label{sec:algorithm}

This section provides the motivation for the AES approach, presents a generic single-population version of it, as well as a multi-population version suitable for ENAS.

\subsection{Overview}
A key problem that AES aims to solve is the inefficiency of synchronous
evaluation strategies when running an EA in a parallel, distributed environment.
This problem is especially challenging for the evolution of DNN architectures that have high variance in
evaluation times due to the various amounts of time needed to train different networks. As a
result, the slowest individuals’ evaluation becomes a bottleneck. This problem can be alleviated through two
mechanisms: (1) If there is a constant supply (i.e.,\ a queue) of individuals to
be readily evaluated, the worker nodes will have optimal throughput and minimal
idle time: They can immediately pull new individuals from the queue
after they have evaluated their current individual. (2) Server idle time can be minimized if
evolution immediately proceeds to the next generation once a certain fraction of
the total number of individuals in the queue has been returned.

Since the number of individuals in the queue exceeds the number of
individuals in a generation, it is not scalable to have the EA server
keep track of all the individuals being evaluated. The solution is simple:
Distribute the bookkeeping to the workers. That is, after the server has placed
the next generation of individuals into the Evaluation Queue, it no longer keeps track of them.
Instead, as workers become available, they pull individuals from the queue, and when they are
done evaluating them, return both the fitness values and the
corresponding individuals back to the server.
Thus, the server only needs to be activated periodically to generate the next generation of individuals.

\begin{figure}[t]
    \begin{center}
        \includegraphics[width=0.75\columnwidth]{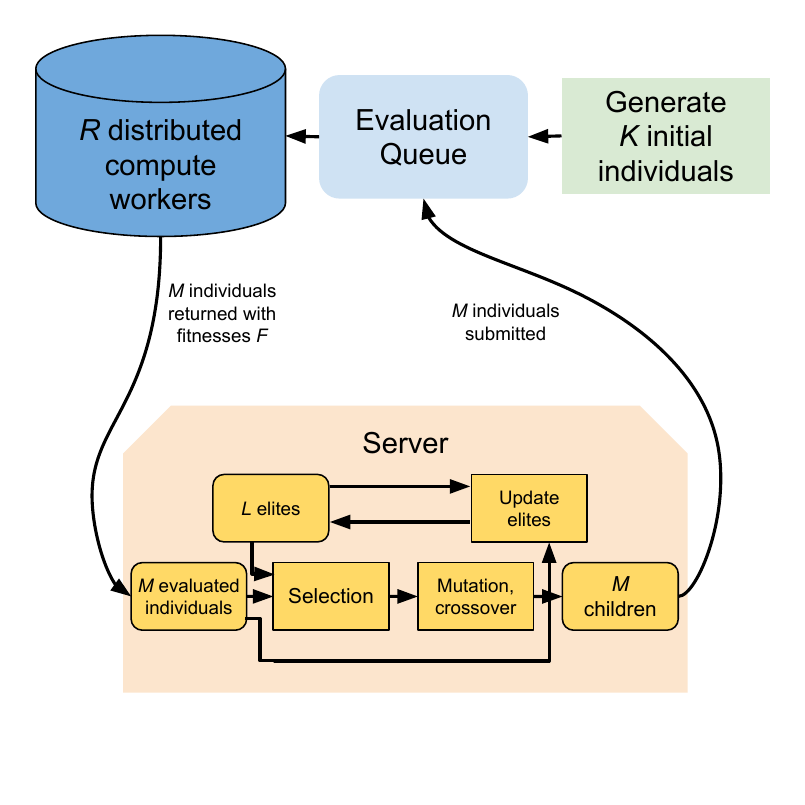}
    \end{center}
    \vspace*{-10ex}    
    \caption{An overview of generic single-population AES. $R$ workers pull individuals from the Evaluation Queue, evaluate them, and return them (with fitnesses $F$) to the server. As soon as $M$ individuals have been returned, the server uses them and the $L$ elite individuals (from the previous generation) to create a new generation of $M$ individuals. They are then placed into the Evaluation Queue, and the elite set is updated. In this manner, the workers in AES do not have to stay idle waiting for a generation to finish their evaluations. The ratio of $M/K=D$ strikes a balance between diversity and efficiency.}
    \label{fg:aesdiagram}
\end{figure}

\subsection{Generic Single-Population AES}

\begin{algorithm}[t]
\caption{Generic Single-population AES}
% \scriptsize
\begin{algorithmic}[1]
    \State Initialize the Evaluation Queue with $K$ individuals (any idle workers will immediately pull individuals from the queue if it is not empty).
    \For {target number of generations}
    \State \parbox[t]{\dimexpr\textwidth-\leftmargin-\labelsep-\labelwidth}{Wait for $M$ individuals and their fitnesses $F$ to return (where $M * D = K$, $D > 1$).\strut}
    \State \parbox[t]{\dimexpr\textwidth-\leftmargin-\labelsep-\labelwidth}{Select $L+M$ parents from the $L$ elites and the $M$ returned individuals.\strut}
    \State \parbox[t]{\dimexpr\textwidth-\leftmargin-\labelsep-\labelwidth}{Generate $M$ child individuals from parents using evolutionary operators.\strut}
    \State \parbox[t]{\dimexpr\textwidth-\leftmargin-\labelsep-\labelwidth}{Update the set of $L$ elites with the $M$ returned individuals.\strut}
    \State \parbox[t]{\dimexpr\textwidth-\leftmargin-\labelsep-\labelwidth}{Submit the $M$ individuals to the Evaluation Queue.\strut}
    \EndFor
\end{algorithmic} 
\label{alg:aes1}
\end{algorithm} 

AES can be easily added onto parallel, synchronous EA with a single population, as shown in Figure~\ref{fg:aesdiagram} and Algorithm~\ref{alg:aes1}. They specify a
generic version of AES with few assumptions regarding the underlying EA
framework and no additional computational burden. The Evaluation Queue is
initialized in the beginning with $K$ randomly generated individuals (Step 1 in Algorithm 1). 
$M$ is the size of the population, i.e.,\ the number of individuals, AES waits to return (Step 3) before it creates the next generation of $M$ individuals. $D$ is a hyperparameter that controls the
ratio between $K$ and $M$. Thus, the number of individuals in the Evaluation Queue decreases from $K-R$ in the beginning of a generation to $K-R-M$ in the end, then jumps back up to $K-R$ as the $M$ newly generated individuals are added to it.

Together, the $M$ individuals that are returned and the $L$ elite individuals from the previous generation constitute the population from which the next generation is created (Steps 4-5).
If any of the returned individuals are better than the worst individuals in the current elite set, the elite set is updated, keeping its size constant at $L$ (Step 6). The entire new generation of $M$ individuals is then submitted for evaluation (Step 7). 
Note that individuals from multiple such generations may be under evaluation at the same time,
so the process is not strictly generational, but can be seen as a mix of generational and steady-state GAs \citep{goswami2023variants}.

As shown in Section~\ref{sec:experiment}, AES can be used to enhance the performance of an existing large-scale EA that utilizes hundreds of thousands of worker nodes.

\begin{figure}[ht]
    \mbox{}\\[-12.5ex]
    \begin{center}
        \includegraphics[width=0.75\columnwidth]{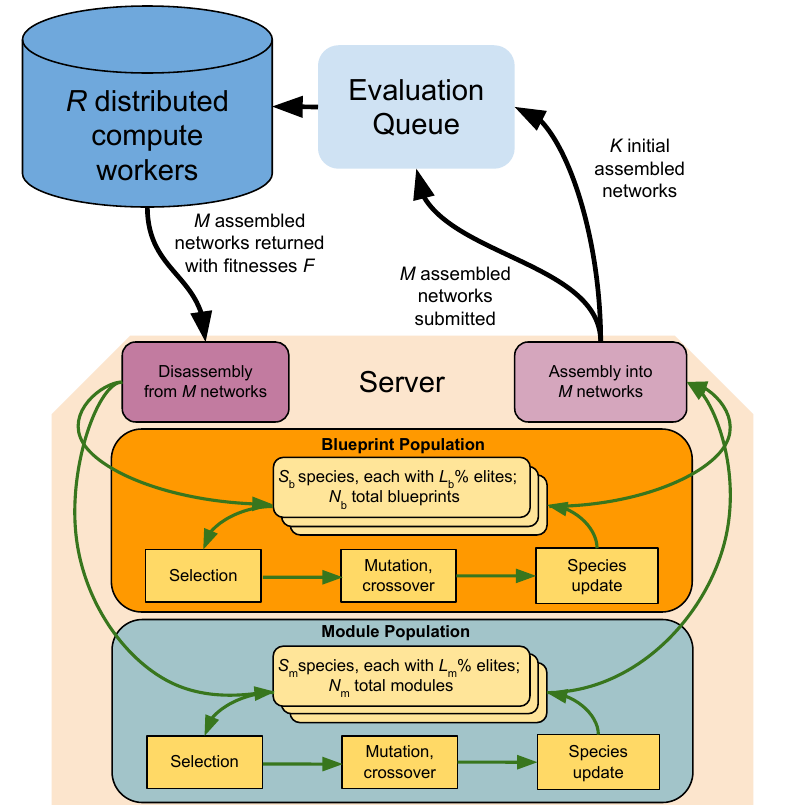}
    \end{center}
    \vspace*{-2ex}    
    \caption{An overview of CoDeepNEAT-AES. The AES process of Figure~\ref{fg:aesdiagram} is extended with blueprint and module populations. Complete networks are assembled from the blueprints and modules and sent to the Evaluation Queue. When $M$ networks are returned, they are disassembled into their blueprints and modules, whose fitnesses are calculated as an average over all networks in which they participated. These blueprints and modules are then merged into the current populations, replacing any existing fitnesses. NEAT neuroevolution is run in each population, replacing the bottom $N-L$ with new blueprints/modules, and updating the species. Similar to generic AES, the workers are fully employed in evaluating individuals, resulting in significant speedup over synchronized CoDeepNEAT.}
    \label{fg:codeepneataesdiagram}
\end{figure}

\subsection{Multi-population AES for ENAS}

\begin{algorithm}[t]
\caption{CoDeepNEAT-AES}
\begin{algorithmic}[1]
    \State Initialize the blueprint population with $N_\mathrm{b}$ random blueprints and divide them into $S_\mathrm{b}$ species.
    \State Initialize the module population with $N_\mathrm{m}$ random modules and divide them into $S_\mathrm{m}$ species.
    \State \parbox[t]{\dimexpr\textwidth-\leftmargin-\labelsep-\labelwidth}{Generate $K$ assembled networks from the blueprint and module populations and submit to the Evaluation Queue.\strut}
    \For {target number of generations }
    \State \parbox[t]{\dimexpr\textwidth-\leftmargin-\labelsep-\labelwidth}{Wait for $M$ assembled networks and their fitnesses $F$ to return (where $M * D = K$, $D > 1$)\strut}
    \State \parbox[t]{\dimexpr\textwidth-\leftmargin-\labelsep-\labelwidth}{Identify all blueprints and modules in the returned networks.\strut}
    \State \parbox[t]{\dimexpr\textwidth-\leftmargin-\labelsep-\labelwidth}{Assign the blueprints and modules the average fitness of their corresponding assembled networks.\strut}
    \State \parbox[t]{\dimexpr\textwidth-\leftmargin-\labelsep-\labelwidth}{Merge the returned blueprints and modules with the existing blueprint and module populations, replacing existing fitnesses.\strut}
    \State \parbox[t]{\dimexpr\textwidth-\leftmargin-\labelsep-\labelwidth}{Keep the top $L_\mathrm{b}\%$ of each blueprint species and $L_\mathrm{m}\%$ of each module species as the elite; discard the rest.\strut}
    \State \parbox[t]{\dimexpr\textwidth-\leftmargin-\labelsep-\labelwidth}{Create new blueprints and modules in each species proportional to their fitness, keeping the population size constant at $N_\mathrm{b}$ and $N_\mathrm{m}$.\strut}
    \State \parbox[t]{\dimexpr\textwidth-\leftmargin-\labelsep-\labelwidth}{Update the species in the blueprint and module populations.\strut}
    \State \parbox[t]{\dimexpr\textwidth-\leftmargin-\labelsep-\labelwidth}{Assemble $M$ new networks from the blueprints and modules.\strut}
    \State \parbox[t]{\dimexpr\textwidth-\leftmargin-\labelsep-\labelwidth}{Submit the $M$ assembled networks to the Evaluation Queue.\strut}
    \EndFor
\end{algorithmic} 
\label{alg:aes2}
\end{algorithm} 

Figure~\ref{fg:codeepneataesdiagram} and Algorithm~\ref{alg:aes2} describe CoDeepNEAT-AES, i.e.,\ a version of AES adapted for the CoDeepNEAT method of ENAS. The
main difference between CoDeepNEAT-AES and the generic AES is
that the evolutionary operations take place at the level of blueprints and modules, not
the evaluated individuals. Therefore, an assembly step is needed before placing individuals
into the Evaluation Queue, and a disassembly step before the evolutionary operations. Also, blueprint and module populations persist continuously across generations: While it would be possible to use only the elites and the returned individuals to construct the population for each generation (as is done in the generic AES), larger and persistent populations provide more comprehensive and diverse source material, making evolution more effective.

There are two populations in CoDeepNEAT-AES: one for blueprints and another for modules. They are both evolved with the NEAT neuroevolution method. One important aspect of NEAT is that it speciates the population automatically, i.e.,\ divides the population into subpopulations of similar individuals, runs evolution primarily within those species (with occasional crossover between species), and adjusts the size of the species according to their overall fitness. In this manner, species may emerge and die out over evolution. 

More specifically, the species that emerge in the module population are numbered and used to supply the modules for each blueprint slot, as shown in Figure~\ref{fg:assembling}.  This assembly step is unchanged in CoDeepNEAT-AES (Step 12 in Algorithm 2).

The disassembly step is more elaborate. It consists of identifying each blueprint and module in each of the returned individuals (Step 6), calculating their fitness as the average of all the networks in which they participated (Step 7), and merging them into the existing populations: If the blueprint or module already exists in the population, the new fitness is used to replace the old one, thus keeping the fitnesses up to date wrt.\ the current other network components (Step 8).

Each population is then evolved as usual with NEAT: Within each species, the top $L\%$ members are preserved as the elite set, and the rest are discarded (Step 9; in CoDeepNEAT-AES, the elite size $L$ is defined as a percentage of the species, instead of an absolute number of individuals as in generic AES). NEAT then assigns new individuals to be generated in each species proportional to their fitness, keeping the total population size is constant at $N$ (Step 10). After these individuals have been generated, the $S$ species are recreated based on the current similarities within the population (Step 11). In this manner, the species change and grow and shrink during evolution.

These extensions make it possible to run AES on a multi-population domain such as CoDeepNEAT ENAS. At the high-level, CoDeepNEAT-AES retains the efficiency of AES, as will be demonstrated experimentally in the next section.

\section{Experimental Results}
\label{sec:experiment}

The generic version of AES was implemented in EC-STAR, a distributed
genetic programming (GP) platform that is scalable to hundreds of thousands of
worker nodes \citep{Shahrzad2015}. It was tested in two single-population domains: Eight-line sorting-network design with known optima, and 11-bit multiplexer design where the goal is to discover a valid solution.  The two experiments serve to evaluate how effective AES is in speeding up evolution when the evaluation times vary a little and when they vary a lot.  Since the evaluation times are relatively short in both domains, it was possible to repeat the runs multiple times and confirm that the differences are statistically significant. For the same reason, it was possible to determine an effective value for the hyperparameter $D$. The third experiment, then, scaled up AES to ENAS, i.e.,\ to CoDeepNEAT neuroevolution in the image-captioning domain. In contrast to the first two experiments, multiple populations are evolved at once, and the optimization is open-ended, i.e.,\ the optimal solution is not known. Also, as is common in deep learning experiments, extremely long evaluation times limit the experiment to comparing single runs. These three experiments thus evaluate AES in two contrasting settings.

\subsection{Sorting-Network Domain}

\begin{figure}[ht]
  \begin{center}
    \includegraphics[width=3in]{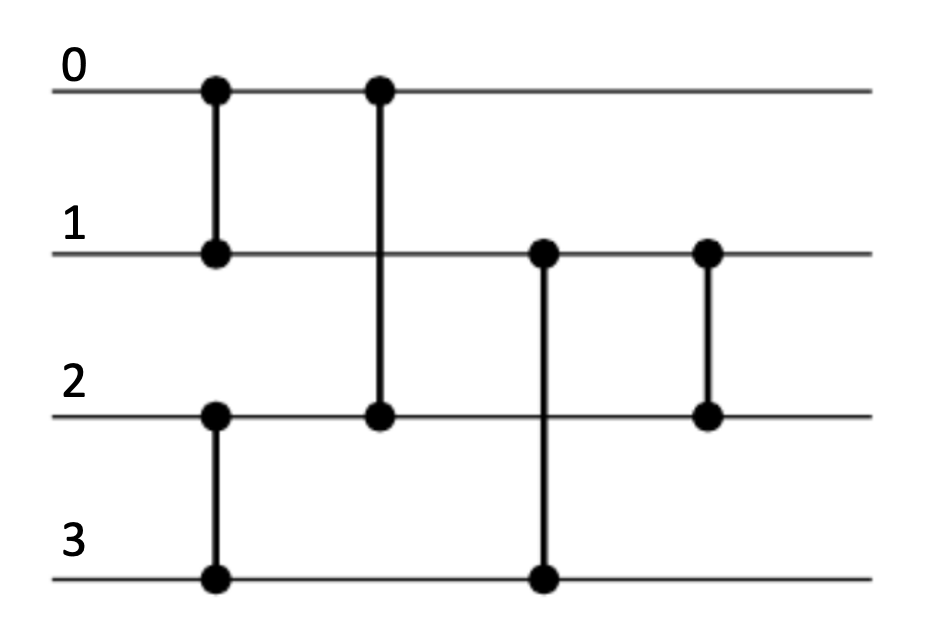}
    \caption{A Four-Input Sorting Network, represented as ((0,1),(2,3),(0,2),(1,3),(1,2)). This network takes as its input (left) four numbers and produces output (right) where those numbers are sorted (large to small, top to bottom). Each comparator (a connection between the lines) swaps the numbers on its two lines if they are not in order, otherwise it does nothing. This network has five comparators, and is the minimal four-input sorting network. Minimal networks are generally not known for input sizes larger than eight, and designing them is a challenging optimization problem.}
    \label{fg:sample-sn}
  \end{center}
\end{figure}

A sorting network of $n$ inputs is a fixed layout of comparison exchange operators (comparators) that sorts all possible inputs \citep[Figure~\ref{fg:sample-sn};][]{knuth:book98}. Since the same layout can sort any input, it represents an oblivious or data-independent sorting algorithm, that is, the layout of comparisons does not depend on the input data. Sorting-network design has been a fundamental problem in computer science for many years, providing an important element for graphics processing units, multi-processor computers, and switching networks \citep{baddar:phd2009, kipfer:hwws04, valsalam:evolint13}. 

Beyond validity, the goal in designing sorting networks is to minimize the number of comparators. Designing such minimal sorting networks is a challenging optimization problem that has been the subject of active research since the 1950s \citep{knuth:book98,shahrzad:alife18,shahrzad:gptp20,valsalam:evolint13}.  For smaller networks of up to eight lines, the optimal solutions are known, making it a verifiable optimization challenge. 

Sorting networks can be represented as a sequence of two-leg comparators where each leg is connected to a different input line and the first leg is connected to a lower line than the second:
\begin{align*}
<\text{\small SortingNetworkConfiguration}> &::=~ <\text{\small ComparatorList}>\\
<\text{\small ComparatorList}> &::=~ <\text{\small Comparator}> {\rm [ <\text{\small ComparatorList}>]}\\
<\text{\small Comparator}> &::=~ <{(f_i, s_i)}>, \text{\small where~} f_i < s_i,~ 0 \leq f_i,s_i < n\\
 \text{\small Example:} & ~((f_1,s_1),(f_2,s_2),...,(f_i,s_i))
\end{align*}
Although the space of possible networks is infinite, it is relatively easy to test whether a particular network is correct: If it sorts all combinations of zeros and ones correctly, it will sort all inputs correctly \citep{knuth:book98}.  This property makes it possible to evaluate networks systematically and efficiently: For instance for an eight-line network, only 256 inputs need to be evaluated to verify that the network is correct.

The evaluation times depend linearly on the size of the network. For instance, evaluating a valid 24-comparator network of eight inputs takes about 25\% more time than evaluating an optimal network with 19 comparators. Such variation in evaluation times is much smaller than with ENAS networks, but provides a test of how much speedup is possible even in a limited case.

\begin{figure}[ht]
  \begin{center}
    \includegraphics[width=\columnwidth]{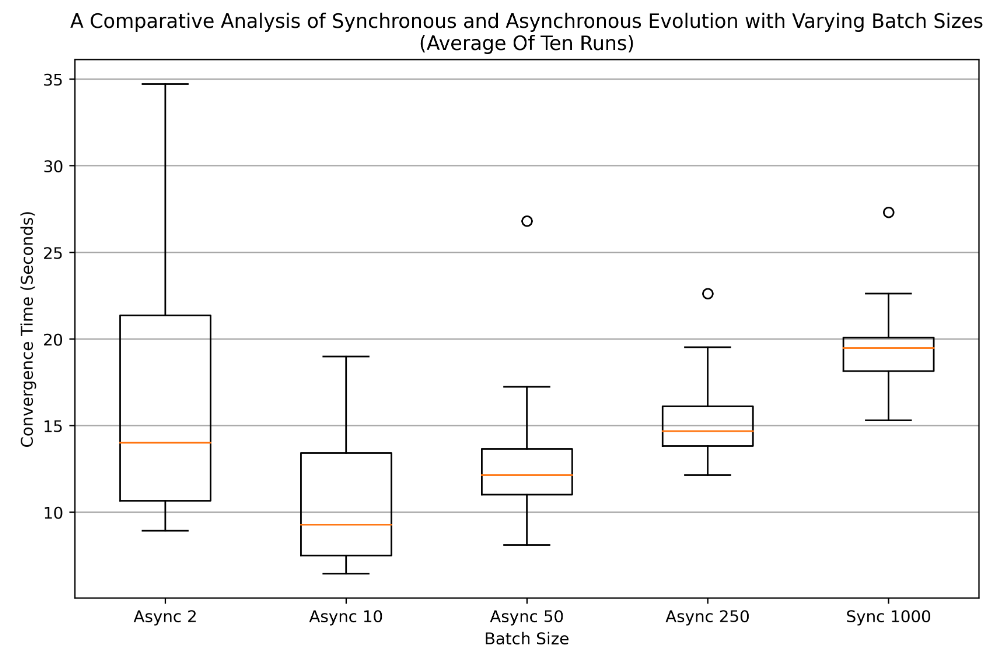}
    \caption{An overview of how different values for $M$ (batch size) affect the convergence time in the sorting-network domain. The settings $K=1000$ and $M=10$ ($D=100$) provide the best performance for this problem.
    The rightmost box ($M$=1000) amounts to synchronous evolution, thus demonstrating that AES results in over two-fold speedup even with limited variation in evaluation times.
    The differences are statistically significant with $p=2.4 \times 10^{-5}$ ($M=10$ vs.\ $M=1000$), $p=4.8 \times 10^{-3}$ ($M=50$ vs.\ $M=1000$), and $p=8.9 \times 10^{-3}$ ($M=250$ vs.\ $M=1000$).}
    \label{fg:sorting}
  \end{center}
\end{figure}

To determine the optimal parameter value for $M$, $K$ was set to 1000, $L$ to 1, and experiments run with $M=$ 2, 10, 50, 250, and 1000. In each experiment, the goal was to find the optimal eight-line sorting network, and the time required on a 32-core machine (i.e.\ $R=32$) was recorded. Each experiment was repeated 10 times.

The results are presented in Figure~\ref{fg:sorting}. The highest speedup is achieved in the midrange, i.e.\ when $M=10$ (i.e.\ $D=100$). In that case, AES finds solutions 2.2 times faster than synchronous evolution (i.e., when $M=K=1000$). This result shows that even with minor variation in evaluation times, AES can achieve significant speedups. What happens with larger variations will be evaluated next.

\subsection{Multiplexer Domain}

Multiplexer functions have long been used to evaluate machine-learning methods
because they are difficult to learn but easy to check \citep{Koza1990}. In
general, the input to the multiplexer function includes $u$ address bits $A_v$
and $2^u$ data bits $D_v$, i.e.,\ it is a string of length $u+2^u$ of the form
$A_{u-1}...A_1 A_0 D_{2^{u-1}}...D_1 D_0$. The value of the multiplexer function
is the value (0 or 1) of the particular data bit that is singled out by the $u$
address bits. For example, for the 11-Multiplexer, where $u=3$, if the three
address bits $A_2A_1A_0$ are 110, the multiplexer singles out data bit
number 6 (i.e.,\ $D_6$) to be its output. A Boolean function with $u+2^u$
arguments has $2^{u+2^u}$ rows in its truth table. Thus, the sample space for the
Boolean multiplexer is of size $2^{u+2^u}$. When $u=3$, the search space is of
size $2^{2^{11}}=2^{2048} \approxeq 10^{616}$ \citep{Koza1990}. However, since
evolution can also generate redundant expressions that are all logically equal,
the real size of the search space can be much larger, depending on the
representation.

Following prior work on the 11-Multiplexer problem \citep{Shahrzad2015}, a rule-based representation was used where each candidate specifies a set of rules of the type
\begin{equation*}
<{\rm rule}> \,\, ::= \,\, <{\rm conditions}> \,\,\rightarrow \,\, <{\rm action}>.
\end{equation*}
The conditions specify
values on the bit string and the action identifies the index of the bit whose
value is then output. For instance, the following rule outputs the value of data
bit 6 when the first three bits are 110:
\begin{equation*}
<A_0=0 \,\, \& \,\, A_1=1 \,\, \& \,\, !A_2=0> \,\, \rightarrow \, D_6.
\end{equation*}
These rules were evolved through the usual genetic operators in genetic programming
\citep{berlanga:infosci10}. Note that with this definition, although logical OR is
not explicitly represented in the grammar, there may be
several rules with the same action. Such a representation is equivalent to a logical OR and allows
the representation to be functionally complete. In other words, the grammar
above, which includes the AND, OR and NOT operators, can be used to express all
possible Boolean functions. This system can produce a range of genes, from only a
single condition rule, up to the maximum number of rules and conditions allowed
per configuration. The maximum number of rules was set to 256 and the maximum number of conditions per rule to 64.

Like the sorting-network domain, the multiplexer domain employs a single population. The search space is much larger, however evolution terminates when a valid solution is found instead of a minimal solution. It is therefore still a simpler setting than ENAS, which includes multiple populations and open-ended optimization.  Also, evaluation times are short enough so that runs can be repeated several times and statistical significance estimated. 
Like ENAS, multiplexer evolution starts with simple solutions and gradually makes them more complex; also, multiplexer solutions require sufficient complexity, as do successful neural networks. The conclusions from the multiplexer are thus likely to carry over to ENAS.

\begin{figure}[ht]
  \begin{center}
    \includegraphics[width=\columnwidth]{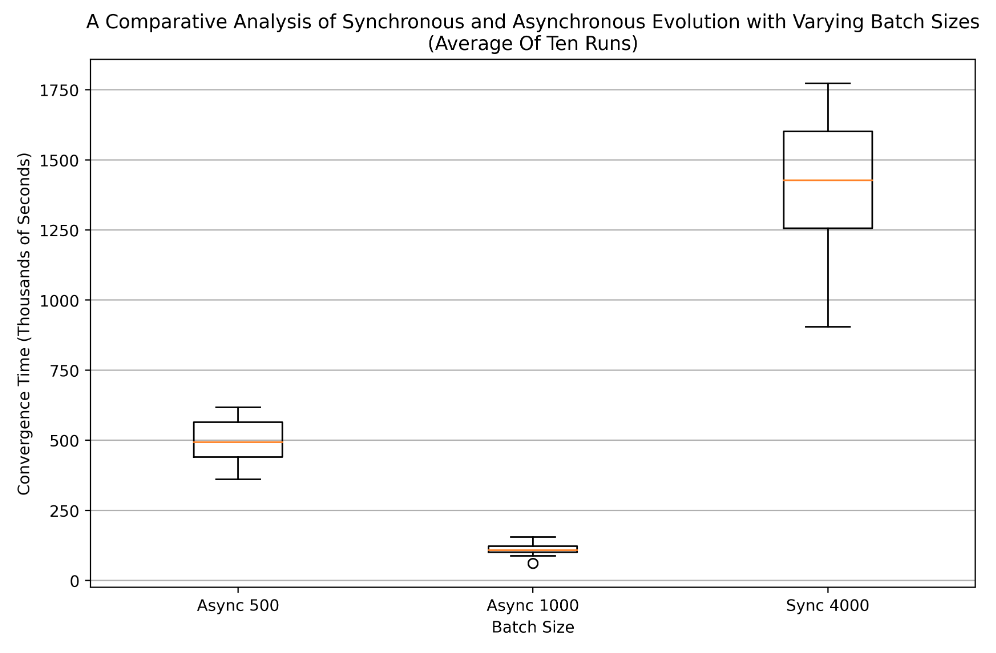}
    \caption{An overview of how different values for $M$ (batch size) affect the convergence time in the multiplexer domain. The settings $K=4000$ and $M=1000$ ($D=4$) provide the best performance for this problem.
    The rightmost box ($M$=4000) amounts to synchronous evolution, thus demonstrating that AES results in a 14-fold speedup in this domain.
    The differences are statistically significant with $p=1.03 \times 10^{-11}$ ($M=1000$ vs.\ $M=4000$) and $p=6.29 \times 10^{-09}$ ($M=1000$ vs.\ $M=500$).}
    \label{fg:multiplexer}
  \end{center}
\end{figure}

However, for the conclusions to carry, it is important to make the evaluation times vary
in the multiplexer the same way as they do in ENAS. In principle, every fitness evaluation
in the multiplexer domain takes a similar amount
of time; therefore, an artificial delay was added to the end of every evaluation. The amount
of delay was modeled after the evaluation timings of an actual run of CoDeepNEAT
on the CIFAR-10 image-classification domain \citep{miikkulainen2019evolving}. Two linear
regression models were fit on a scatter plot of (1) the mean evaluation time vs.\ the number
of generations elapsed, and (2) the standard deviation of evaluation time vs.\ the
number of generations elapsed. During each generation of EC-Star, the
two linear models were used to predict the mean and standard deviation;
these values were used to construct a Gaussian distribution from which the delays
for fitness evaluations were sampled.

In order to determine an appropriate value for $M$, $K$ was set to
4000, $L$ to 100, and three different values of $M$ tested (500, 1000, 4000). In each test, the amount of
time necessary for EC-Star to converge and solve the multiplexer problem was recorded, using $R=4000$.
The experiments were repeated 10 times for each value of $M$.

The results are summarized in Figure~\ref{fg:multiplexer}, which plots
convergence time for the different $M$ values. Interestingly, setting $M$ to an extremely
low or high value can hurt performance. Too small batches are akin to too small populations: Enough diversity is needed in the batch to allow evolution to progress well.  On the other hand, too large batches result in longer evolution. In cases where $M=1000$, evolution shows
the most substantial speedups. In this case $D$ is approximately 4, in contrast with $D=100$ in the sorting network domain. Thus, depending on the amount of variation in the evaluation times, $D$ can be adjusted to obtain significant speedups. In the multiplexer domain, AES finds solutions 14 times faster than synchronous evolution (i.e.,\ when $M=K=4000$), which is a remarkable speedup indeed.

\subsection{Image-Captioning Domain}

Deep learning has recently provided state-of-the-art performance in image
captioning, and several diverse architectures have been suggested
\citep{vinyals:cvpr15,xu:icml15,karpathy:cvpr15,you:cvpr16,vedantam:arxiv17,hossain2019comprehensive,polap2023hybrid}.
The input to an image-captioning system is a raw image, and the output is a text
caption describing the contents of the image. In many of these
architectures, a convolutional network is used to process the image into
an embedding. This embedding is then given to recurrent layers such as LSTMs to
generate coherent sentences with long-range dependencies. Further, an U-NET convolutional architecture can be used to segment objects before classifying them for captioning, improving performance \citep{polap2023hybrid}. 

As is common in existing approaches, a pretrained ImageNet model
\citep{szegedy:cvpr16} was used to produce the initial image embeddings. The evolved
network took an image embedding as input, along with a sequence of one-hot text
inputs. During training, the text input contained the previous word of the ground
truth caption; during inference, it contained the previous word generated by the model
\citep{vinyals:cvpr15, karpathy:cvpr15}. In the initial CoDeepNEAT population, the
image and text inputs were fed to a shared embedding layer, which was densely
connected to a softmax output over words. From this simple starting point,
CoDeepNEAT evolved architectures that included fully connected layers, LSTM
layers, pooling layers, concatenation layers, as well as sets of hyperparameters associated
with each layer, along with a set of global hyperparameters \citep{miikkulainen2019evolving}.
In fact, the well-known Show-and-Tell image-captioning architecture
\citep{vinyals:cvpr15} is in this search space.

Two separate runs of CoDeepNEAT for evolving DNNs in the
image-captioning domain were performed. The baseline version of CoDeepNEAT was synchronous, while the improved version, called CoDeepNEAT-AES, made use of asynchronous evaluations. To keep the
computational costs reasonable, during evolution, the networks were trained for
six epochs, and on one-fifth of the entire MSCOCO image-captioning
dataset. Identical hyperparameters were used in both runs: Population sizes were $N_\mathrm{b}=20$ and $N_\mathrm{m}=60$, $L_\mathrm{b}=L_\mathrm{m}=50\%$, divided into $S_\mathrm{b}=1$ and $S_\mathrm{m}=3$ species.
For CoDeepNEAT-AES, $K=300$ and $M=100$ (i.e.,\ $D=3$, on par with $D=4$ in the multiplexer experiment) were used. The worker nodes were composed
of up to $R=200$ Amazon EC2 spot instances with GPU support for training DNNs.
%and the completion service provided the interface between them and the server. 
% Due to cost concerns of running so many EC2 instances, a value of $D=3$ is used.
Because EC2 spot instances are inherently unreliable and may be temporarily
unavailable for any reason, both runs were started at the same time to
remove a potential source of bias. Each method was run until convergence,
which took about 89 hrs. Due to this cost (in terms of time, money, and carbon footprint)
the conclusions were drawn from these single runs, as is common in modern
deep-learning experiments.

\begin{figure}[ht]
  \begin{center}
    \includegraphics[width=\columnwidth]{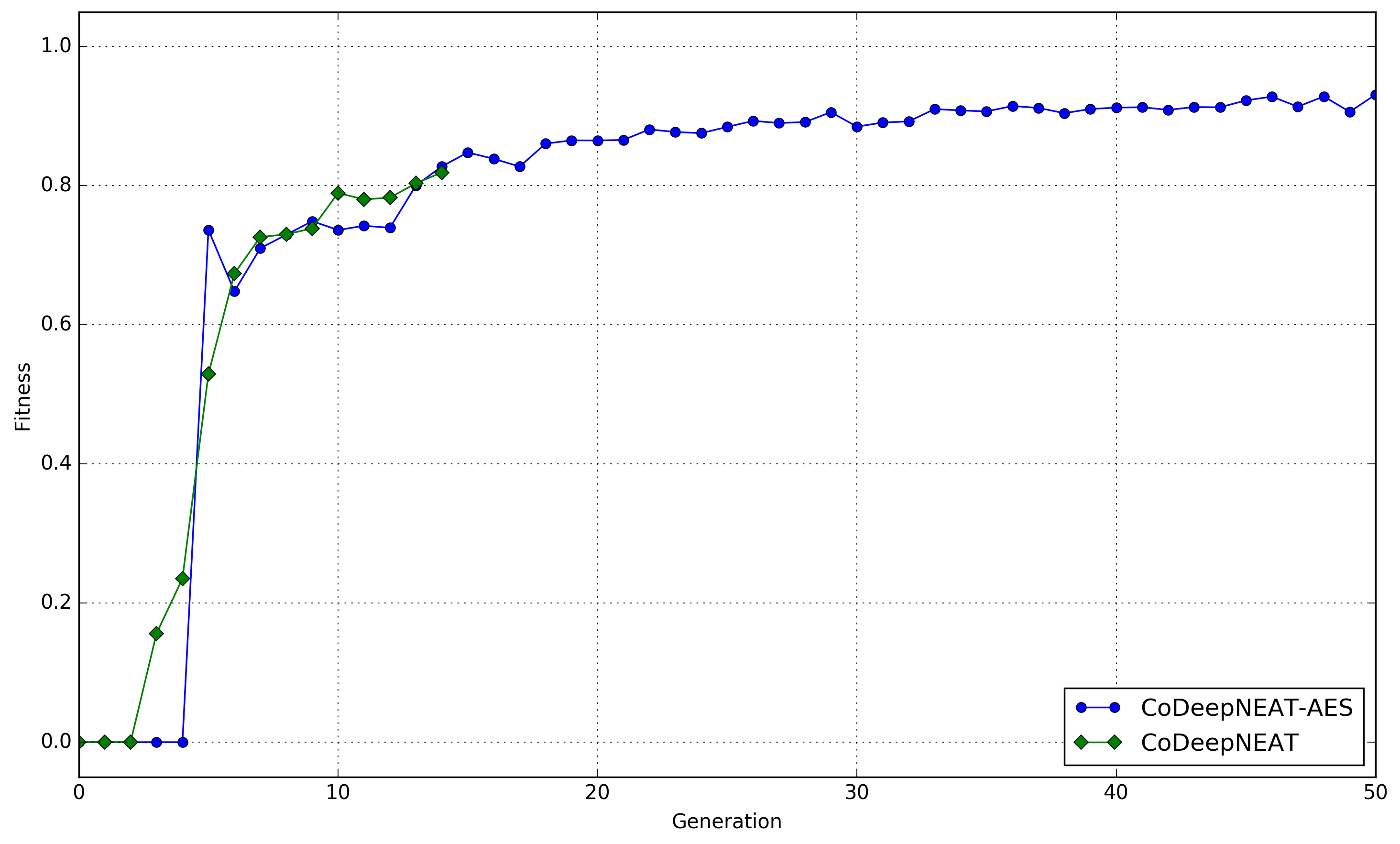}
    \caption{A plot of fitness vs.\ number of generations elapsed for synchronous CoDeepNEAT and CoDeepNEAT-AES. The algorithms perform comparably at each generation. However, CoDeepNEAT-AES is much faster, as seen in Figures~\ref{fg:codeepneat_gen} and~\ref{fg:codeepneat_time} (for this reason, the CoDeepNEAT was run only 14 generations while CoDeepNEAT-AES was run until 50.)}
    \label{fg:codeepneat_gen}
  \end{center}
\end{figure}

\begin{figure}[ht]
  \begin{center}
    \includegraphics[width=\columnwidth]{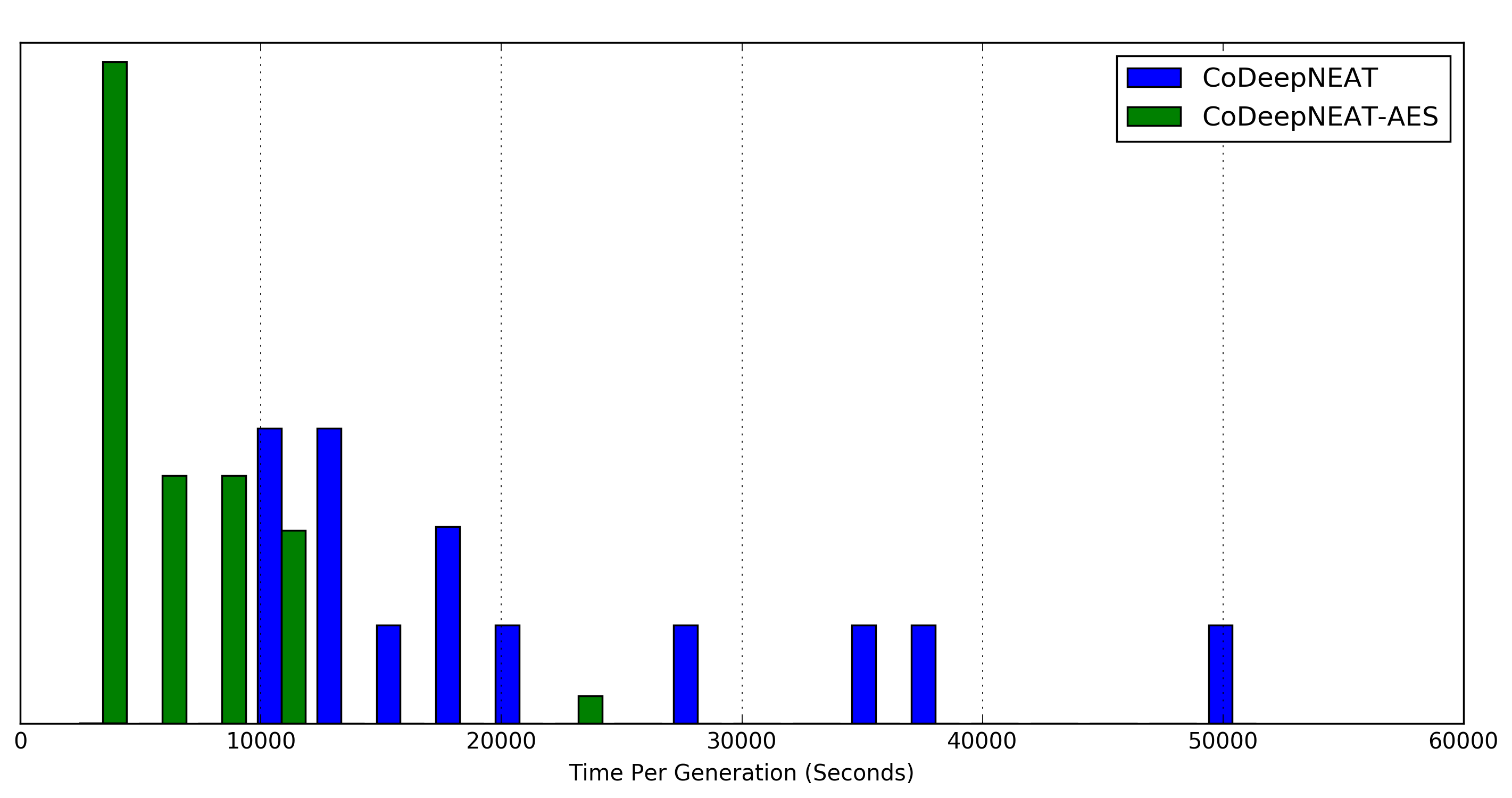}
    \caption{A histogram of time per generation for synchronous CoDeepNEAT and CoDeepNEAT-AES. CoDeepNEAT-AES uses significantly less time per generation.
%    in the worst case scenario.
}
    \label{fg:codeepneat2_top}
  \end{center}
\end{figure}

\begin{figure}[ht]
  \begin{center}
    \includegraphics[width=\columnwidth]{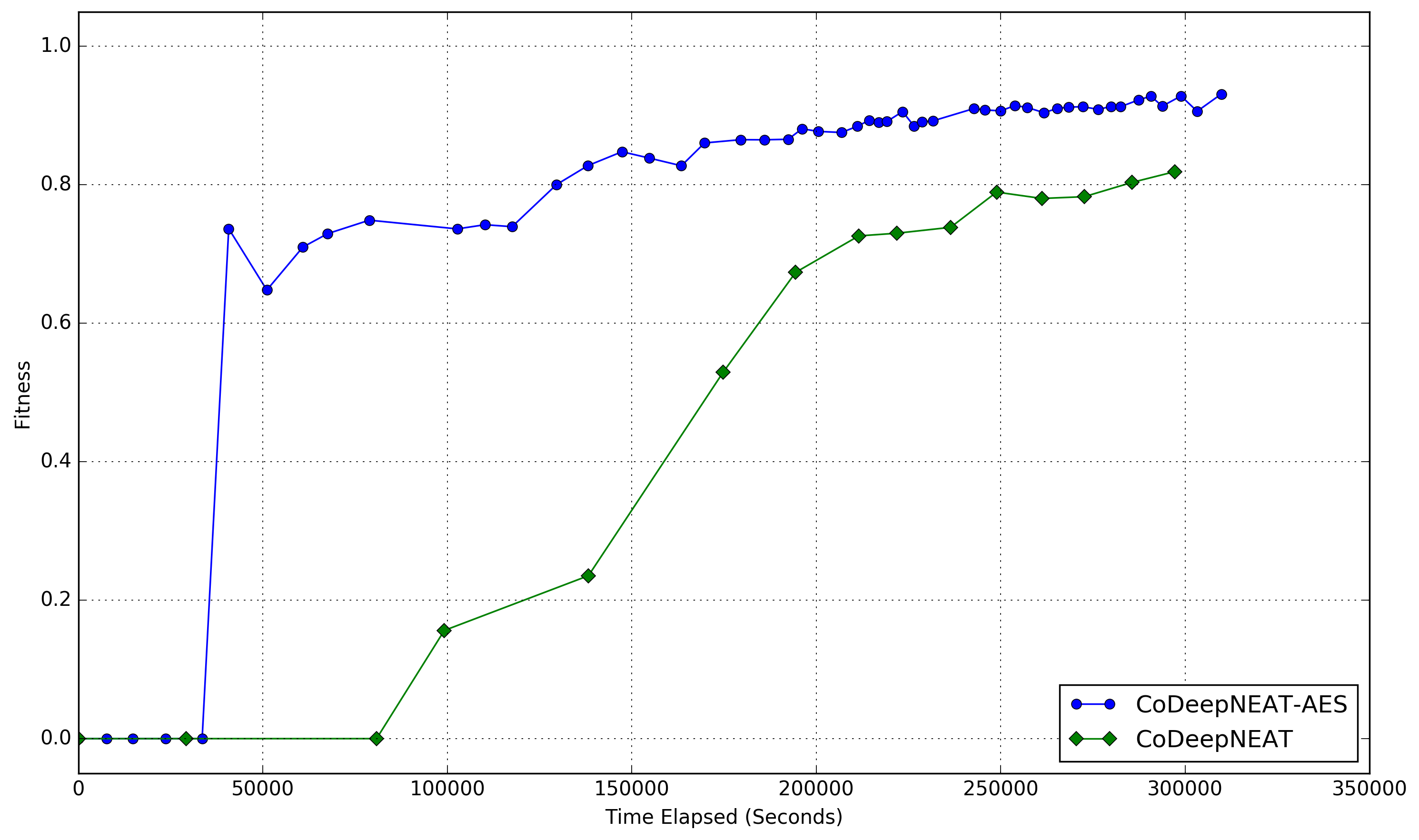}
    \caption{A plot of fitness vs.\ wallclock time elapsed for synchronous CoDeepNEAT and CoDeepNEAT-AES. Each marker in the plot represents the fitness at a different generation. CoDeepNEAT-AES improves faster than regular CoDeepNEAT and achieves a higher fitness in the same amount of time.}
    \label{fg:codeepneat_time}
  \end{center}
\end{figure}

The CoDeepNEAT and CoDeepNEAT-AES runs resulted in a similar range of architectures, similar to those reported in the original CoDeepNEAT experiments \citep{miikkulainen2019evolving}.
From Figures~\ref{fg:codeepneat_gen},~\ref{fg:codeepneat2_top} and~\ref{fg:codeepneat_time}, it is clear that CoDeepNEAT-AES runs significantly faster than the synchronous version of CoDeepNEAT. Although both versions achieve similar fitness after the same number of generations (Figure~\ref{fg:codeepneat_gen}), each generation of CoDeepNEAT-AES takes far less time (Figure~\ref{fg:codeepneat2_top}), and as a result, CoDeepNEAT-AES progresses much faster in wall-clock time (Figure~\ref{fg:codeepneat_time}). At 130,000 seconds elapsed, CoDeepNEAT-AES was able to reach the same fitness as CoDeepNEAT at 300,000 seconds, thus resulting in a 2.3-fold speedup. Also, when run until 300,000 seconds, CoDeepNEAT-AES was able to find better solutions than CoDeepNEAT.
Overall, the experimental results suggest that AES accelerates the convergence of CoDeepNEAT over two fold in the image-captioning domain.

\section{Discussion and Future Work}
\label{sec:discussion}

As the experimental results show, AES provides a significant speedup in
the sorting-network, multiplexer, and image-captioning domains. Furthermore, the hyperparameter
$D$ is crucial to the improved performance of AES. With little variation in the evaluation times, it needs to be large; with more variation, small. When $D$ approaches 1 (i.e.,\ $M$ approaches $K$)
AES becomes similar to synchronous evaluation, and increasingly needs to wait for the slowest individuals to finish evaluation.
However, setting a value for $D$ that is too large also hurts
performance. This is likely because as $M$ gets smaller, both the returned
individuals and the new population that is generated from them become less
diverse.

\begin{figure}[ht]
  \begin{center}
    \includegraphics[width=\columnwidth]{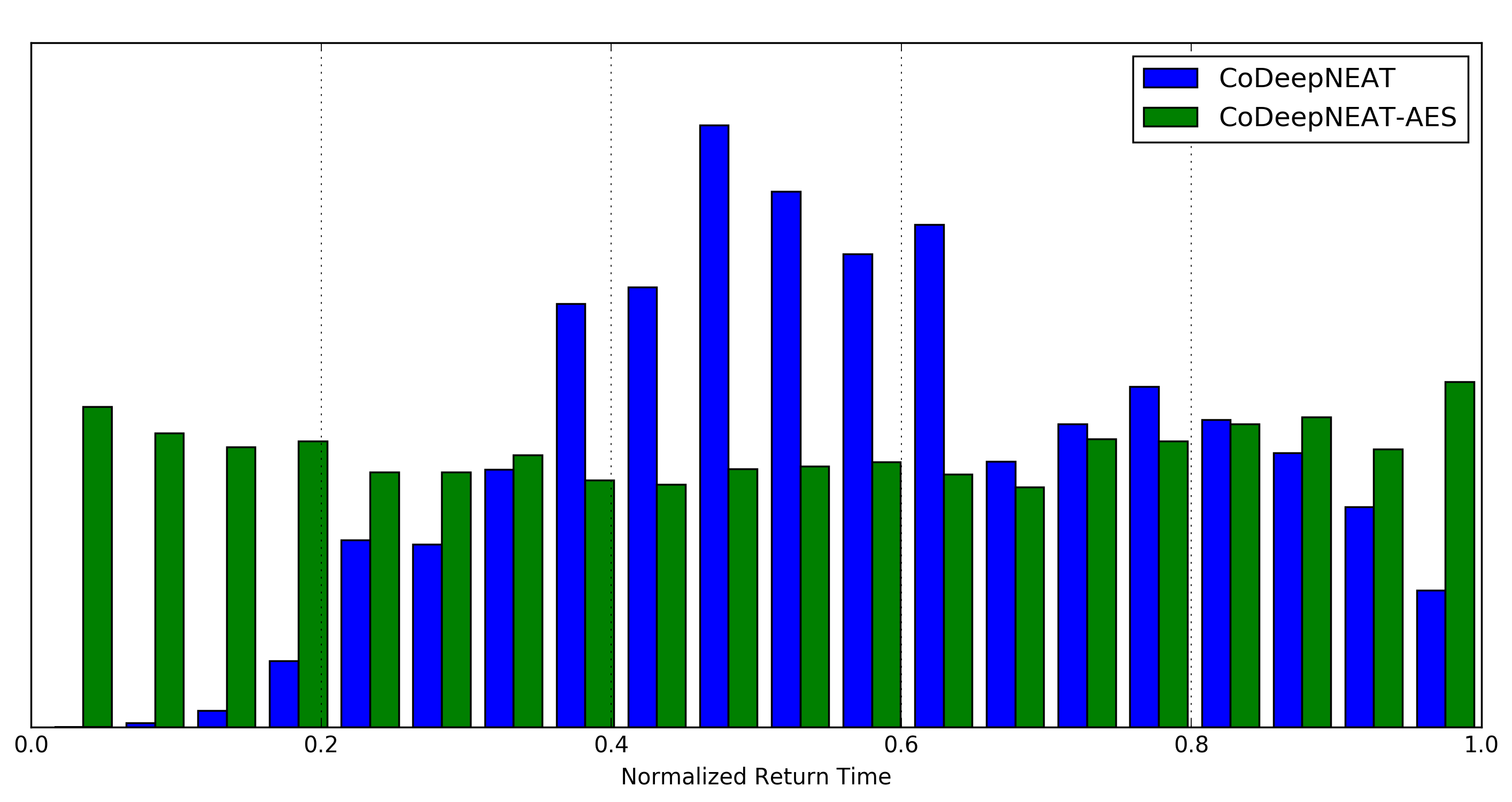}
    \caption{A histogram of the times when individuals return from evaluation over the course of an average generation for both algorithms. CoDeepNEAT-AES has a uniform distribution while CoDeepNEAT has a Gaussian distribution, thus demonstrating that CoDeepNEAT-AES wastes little time waiting for slow individuals.} 
    \label{fg:codeepneat2_middle}
  \end{center}
\end{figure}

The histogram in Figure~\ref{fg:codeepneat2_middle} reveals how AES improves
performance over synchronous evaluation. This plot visualizes the
distribution of times at which individuals (along with their fitness) return from
evaluation over the duration of an average generation. In the
synchronous version of CoDeepNEAT, individuals in the population are submitted
at the same time, and all come back in the same generation before evolution can proceed. As a
result, the histogram for synchronous CoDeepNEAT is roughly a Gaussian distribution,
with some individuals returning early and some returning late. A lot of time is thus
wasted waiting for the last few individuals. On the other hand, this delay does not occur with CoDeepNEAT-AES; the
distribution is uniform, indicating that individuals are returned at a
steady, regular rate over the course of a generation and there are no
slow individuals that might bottleneck the EA.

\begin{figure}[ht]
  \begin{center}
    \includegraphics[width=\columnwidth]{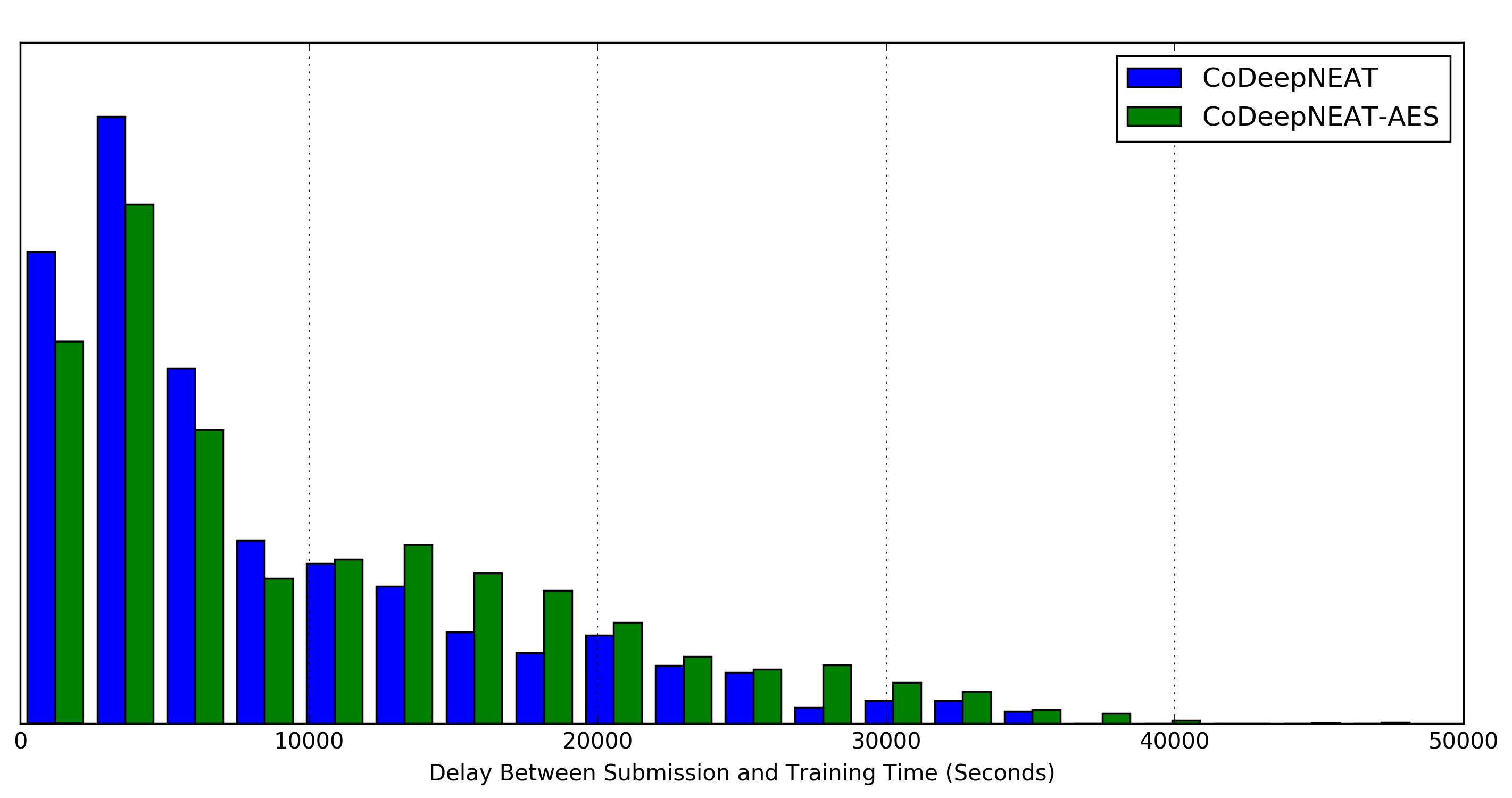}
    \caption{A histogram comparing the delay between submission of individuals to the Evaluation Queue and when they actually start training. CoDeepNEAT-AES has a slightly higher delay, but it is not sufficient to affect performance (as demonstrated by Figure~\ref{fg:codeepneat_time}).}
    \label{fg:codeepneat2_bottom}
  \end{center}
\end{figure}

There is one measure where the synchronous version of CoDeepNEAT has
an advantage. This result is seen in the histogram in
Figure~\ref{fg:codeepneat2_bottom}, which visualizes the delay between
when an individual is placed in the Evaluation Queue and when
that same individual assigned to a worker node. The delay is
slightly higher for CoDeepNEAT-AES than the synchronous version of CoDeepNEAT. The reason is 
that CoDeepNEAT-AES maintains more individuals in the Evaluation Queue. However, as the fitness plot in Figure~\ref{fg:codeepneat_time}
indicates, this longer delay does not affect performance significantly. 

Another aspect of CoDeepNEAT-AES that differs from its synchronous counterpart is that it tends to favor candidates with lower evaluation times: Their lineage can go through more generations in the same amount of time. Evidence of this effect can be seen in Figure~\ref{fg:codeepneat2_top} where the mean amount of time spent per generation is significantly lower for CoDeepNEAT-AES than for CoDeepNEAT. While asynchronous EAs are known to have such evaluation bias, and efforts have been developed to avoid it \citep{guijt2023impact}, it is not undesirable in the case of neuroevolution. Discovering DNNs that are faster to train is often a secondary goal of many architecture search algorithms, and as Figures~\ref{fg:codeepneat_gen} and~\ref{fg:codeepneat_time} show, CoDeepNEAT-AES is able to achieve the same quality of solutions as CoDeepNEAT while taking much less time. 

Although AES was developed primarily as a method for ENAS, it is a general method of asynchronous evolution. In future work, CoDeepNEAT-AES may be combined with other
improvements such as age-layering and learning curve prediction \citep{hodjat:alife16,klein2017learning}. Furthermore,
more extensive experiments can be done to analyze how different values for $K$
and $D$ will affect the performance of CoDeepNEAT-AES.  It should also be possible to use the generic version of AES to scale up evolutionary experiments in many other domains as well.

\section{Conclusion}
\label{sec:conclusion}

This paper proposed a new asynchronous EA called AES designed for complex problems such as optimizing the architecture of DNNs. It can use the available distributed computing resources efficiently with both single and multi-population EAs, and in verifiable discovery and in open-ended optimization tasks. AES works by maintaining a queue of networks that are ready to be evaluated, and by proceeding with evolution once a fraction of the networks have returned from the workers. Experimental results in the sorting-network, multiplexer, and image-captioning domains show that AES can attain a two to 14-fold speedup over its synchronous counterpart with no loss in accuracy or final fitness. AES is thus a promising way to extend evolutionary optimization to complex domains where traditional parallelization methods are ineffective. 

%% If you have bibdatabase file and want bibtex to generate the
%% bibitems, please use
%%
\bibliographystyle{elsarticle-harv} 
\bibliography{nnstrings,nn,risto,img_cap}

\end{document}